\documentclass[letterpaper, 10 pt, conference]{ieeeconf}      % Use this line for conference
\usepackage{amssymb,graphicx,amsmath,color,algorithm,algorithmic,url,blindtext}%,tabularx,subcaption,float}
\usepackage[english]{babel}
\usepackage{color,soul}
\usepackage{multirow}
\usepackage{anyfontsize} %For Subtitle in Python
\usepackage{placeins}
\usepackage{subcaption}
\usepackage{hyperref}
\usepackage[noadjust]{cite} % groups citations
\usepackage{hhline,caption}
\IEEEoverridecommandlockouts                              % This command is only
\title{
Uncertainty-aware deep learning for robot touch:\\
Application to Bayesian tactile servo control
}

\author{Manuel Floriano V\'{a}zquez, Nathan F. Lepora%
\thanks{This work was supported by an award from the Leverhulme Trust to NL on `A biomimetic forebrain for robot touch' (RL-2016-39)}
\thanks{MFV is a MSc Robotics postgraduate student from University of Bristol. \newline
Email: yr19556@bristol.ac.uk}
\thanks{NL is with the Department of Engineering Mathematics and Bristol Robotics Laboratory, University of Bristol, Bristol, U.K. \newline
Email: n.lepora@bristol.ac.uk}
}
\begin{document}

\maketitle

%%%%%%%%%%%%%%%%%%%%%%%%%%%%%%%%%%%%%%%%%%%%%%%%%%%%%%%%%%%%%%%%%%%%%%%%%%%%%%%%
\begin{abstract}
This work investigates uncertainty-aware deep learning (DL) in tactile robotics based on a general framework introduced recently for robot vision. For a test scenario, we consider optical tactile sensing in combination with DL to estimate the edge pose as a feedback signal to servo around various 2D test objects. We demonstrate that uncertainty-aware DL can improve the pose estimation over deterministic DL methods. The system estimates the uncertainty associated with each prediction, which is used along with temporal coherency to improve the predictions via a Kalman filter, and hence improve the tactile servo control. The robot is able to robustly follow all of the presented contour shapes to reduce not only the error by a factor of two but also smooth the trajectory from the undesired noisy behaviour caused by previous deterministic networks. In our view, as the field of tactile robotics matures in its use of DL, the estimation of uncertainty will become a key component in the control of physically interactive tasks in complex environments. 
\end{abstract}

%Keywords appear just beneath the abstract. Use only for final RAL version.
%\begin{IEEEkeywords} Force and Tactile Sensing; Biomimetics \end{IEEEkeywords}

%%%%%%%%%%%%%%%%%%%%%%%%%%%%%%%%%%%%%%%%%%%%%%%%%%%%%%%%%%%%%%%%%%%%%%%%%%%%%%%%
\section{INTRODUCTION}

% Why the sense of touch? What is tactile robotics?
%Typically, the sense of touch in humans is used to characterize shapes or surfaces via the shearing of our skin and the pressure exerted \cite{pohl2012engaging}, which enables us to interact dexterously with our physical environment. 
Motivated by a growing interest in robots that can interact with complex environments, the field of tactile robotics seeks to endow machines with the capabilities to sense the intricacies of physical contact so that interaction may be controlled.
% Difficulties in its implementation
%Tactile robotics has some similarities with Computer Vision (CV), e.g. a renewed impetus from the Deep Learning (DL) revolution that is now impacting robot vision. However, 
Tactile robotics differs from computer and robot vision in important ways. The need for physical contact to deform the sensing surface introduces a much more complex data analysis problem. This complexity is further compounded by there being many different designs of tactile sensor that can be expensive or difficult to deploy. Overall, this has resulted in a slow development of the field of robot touch~\cite{lepora2020optimal}.

% Why focus on uncertainty
Meanwhile, far from being perfect, model-based vision using deep learning in robotics is still facing complex problems such as the ability to deal with uncertain information~\cite{sunderhauf2018limits}. As the behaviour of vision-based robots depends largely on their perception, deterministic approaches cannot be blindly trusted. Consequently, there is interest in developing uncertainty-aware deep learning (DL) networks that are able to self-estimate a degree of belief in their predictions.

%which can be later combined with Bayesian filtering over these predictions to improve the quality of the perception.

\begin{figure}[htb]
    \centering
    \includegraphics[width = \columnwidth]{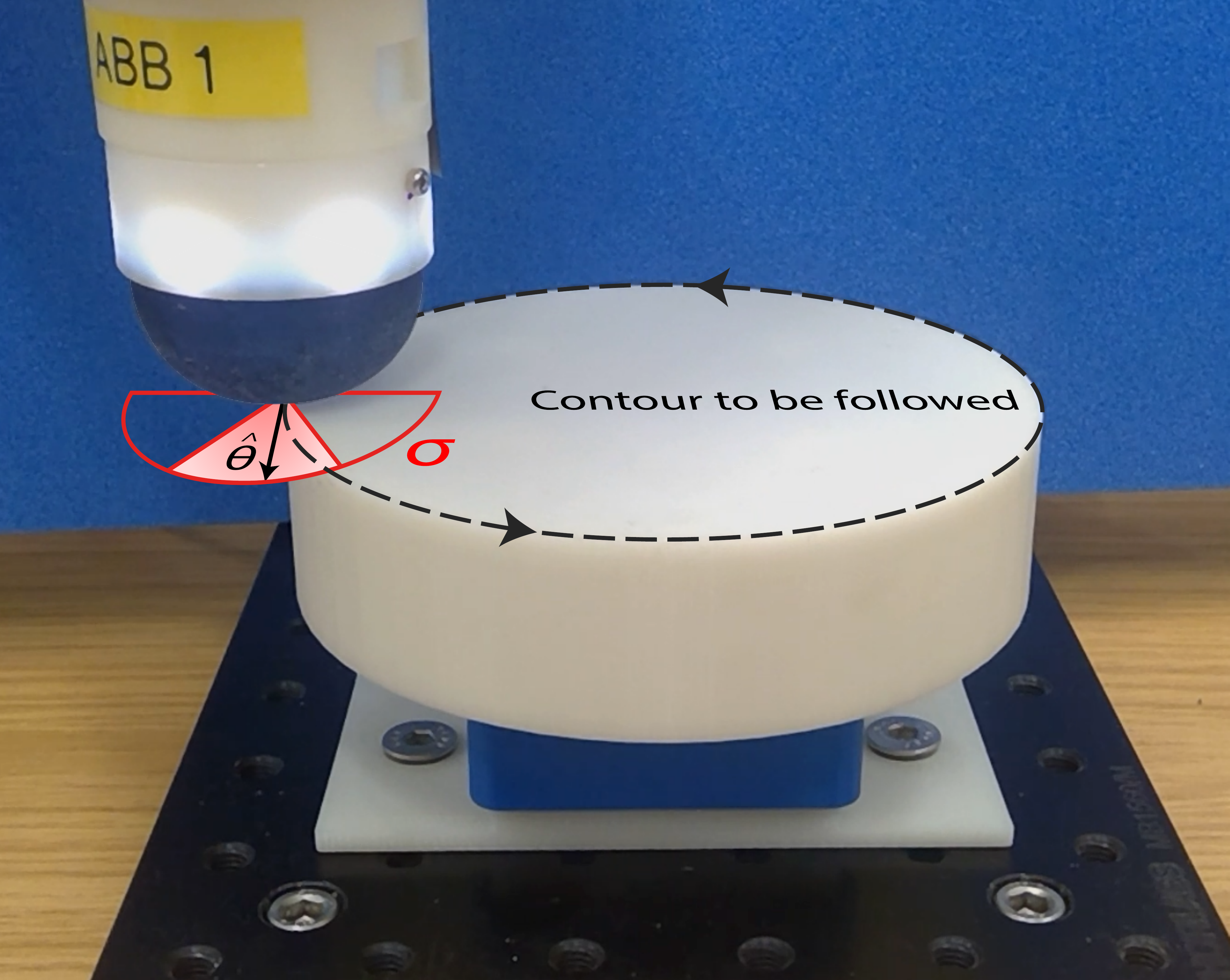}
    \caption{Tactile servo control task around a disk. Estimates of the edge pose for controlling the sensor are subject to noise, such as from sensor shear. Awareness of the uncertainty can improve the control by using Bayesian filtering methods.}
    \label{fig:ABB}
\end{figure}

% What I do and What can I contribute with
In this work, we make a first investigation of uncertainty-aware DL in tactile robotics based on a general framework for uncertainty estimation in DL that was recently introduced for robot vision \cite{loquercio2020general}. We examine uncertainty-aware tactile DL both offline on pre-collected tactile data from moving against a test object and online in performing a tactile servo control task to follow a contour around an object (\autoref{fig:ABB}). Servo control is considered because it is representative of tasks controlled by robot touch that we expect to have broader applicability in the control of dexterous robots. Our hypothesis is that uncertainty-aware DL combined with Bayesian filtering will improve the predictions, leading to better tactile servo control.  

%Traditionally, tactile sensors were implemented as arrays of discrete tactile elements (taxels or tactels) covering the sensing surface \cite{iwata2009design, funabashi2018versatile}, which generates relatively low-dimensional outputs typically around 10-50 over a fingertip area. Recent developments have favoured th

% Optical tactile sensors and its affinity with DL. Remind that optical tactile sensors exploits the progress made in Robotic Vision and CV. 
This investigation is based on the use of optical tactile sensors capable of generating high-dimensional information, which have shown a great affinity to techniques developed for computer vision such as convolutional neural networks~\cite{abad2020visuotactile}. Optical tactile sensors that capture the deformation of the sensing surface with an internal camera include the GelSight \cite{johnson2009retrographic,yuan2017gelsight} from MIT and the biomimetic TacTip sensor \cite{chorley2009development,ward2018tactip} from BRL that is used in this research. Recent studies have shown how the TacTip's use of pin/papillae deflections under its sensing surface in combination with deep learning enables accurate estimation of the local pose of object features~\cite{lepora2020optimal}, which can then be used to control the tactile sensor to follow the contours of complex objects~\cite{lepora2019pixels}. Here we demonstrate that uncertainty-aware DL can improve the pose estimation over the previous state of the art. The robot is then able to robustly follow all of the presented contour shapes to reduce not only the error but also the undesired noisy behaviour caused by previous deterministic networks. 

%% Delete predicted contour; replace 'real contour' with 'contour to be followed' (and put an anticlockwise arrows on the contour to indicate motion; I think the uncertainty representation will be improved by adapting the semicircle with a red section from Loquercio et al 2020 (Fig 1) works well - this could be tilted to an semi-ellipse to indicate the uncertainty in the edge orientation perception

% To choose between tactile_uncert2.png and tactile_uncert.png

% Content review map

\clearpage
\section{BACKGROUND AND RELATED WORK}

\subsection{Deep Learning with Optical Tactile Sensing}

% Affinity Optical tactile sensors - DL
The capacity of deep learning to address problems involving high-dimensional data has shown promise for tactile robotics since it was first introduced for discrete taxel-based sensors in 2014 \cite{schmitz2014tactile} and then first applied to optical tactile sensing in 2017 with the GelSight  \cite{yuan2017shape}.
% Change to DL models in TacTip
The successes of DL with the GelSight \cite{yuan2017shape,Yuan2017b,calandra2017feeling,Calandra2018,Hogan2018,tian2019manipulation,luo2018vitac,Lee2019} motivated the use of convolutional neural network (CNN) models for the TacTip optical tactile sensor considered here~\cite{lepora2019pixels,lepora2020optimal,james2019}, with a particular focus on the closed-loop control of the sensor as it interacts with complex objects. The improved perception system led to a step-change in the robustness and accuracy of the control compared to previous machine learning algorithms that estimated likelihoods of discrete perceptual classes from time series of pin displacements extracted from the tactile images with standard CV techniques \cite{ward2018tactip,lepora2017exploratory}. As a result, it was observed that CNN model led to advantages including the ability to: 1) isolate and remove undesired shear effects as the soft curved sensor moves against the object surface, and 2) generalize over unseen situations such as curved surfaces~\cite{lepora2019pixels,lepora2020optimal}. The implementation of this approach uses the CNN as input to a PI controller to robustly follow different contours and shapes in 3D \cite{lepora2020optimal}. However, as these algorithms rely on deterministic CNNs for tactile perception, a single prediction can greatly affect the response of the controller, introducing an unnecessary corrective behaviour and thus a more inaccurate trajectory.

\subsection{Uncertainty estimation in Robotic Vision}
%Why uncertainty?
Existing approaches for uncertainty estimation seek to characterize two different types of sources: those produced by the sensor noise (data uncertainty) and those derived from the biases of the trained network (model uncertainty).
% Noisy sensors (BNN)

Data uncertainty is estimated  typically by replacing intermediate activation functions by probability distributions via the method of Assumed Density Filtering (ADF) \cite{boyen2013tractable}, giving rise to Bayesian Neural Network (BNN) that was first implemented in Ref.~ \cite{frey1999variational} and later optimized~\cite{gast2018lightweight}. These architectures assume Gaussian distributions to forward propagate the noise of the sensor to the output.

% MC Sampling (Model uncertainty)
Model uncertainty depends on how many training samples have been used and their distribution. Intuitively, those input samples fed into the network that differ from the training samples should be associated with higher uncertainty. This process has been modelled with specialized architectures \cite{lee2019ensemble, kahn2018self, lakshminarayanan2017simple} or the use of Monte Carlo (MC) sampling with dropout enabled during testing \cite{gal2016dropout}. The latter approach estimates the posterior probability distribution by randomly disabling some neurons and repeatedly feeding the same input. By doing so, an unseen sample introduced into the network will produce a higher output variance as a consequence of the lack of redundancy acquired by the architecture. Thus, the model uncertainty can be inferred by properly tuning the dropout and the number of samples used.

% General uncertainty framework
Recently, Loquercio {\em et al} introduced a general uncertainty framework \cite{loquercio2020general} that uses MC sampling in BNNs with dropout enabled during testing, achieving a superior performance over state-of-the-art methods. Their approach not only estimates the data and model uncertainty, but also models the relationship between these sources of uncertainty.
% How this related to our work. Why did we choose it?
Given those results and the release of an open-source toolbox for implementing their framework, we have chosen to adopt those methods for the uncertainty estimation of the tactile perception system in the present work. %These methods are used to filtering to improve the quality of the predictions and thus providing a robust sliding motion.

\section{METHODS}

\subsection{Training the deterministic CNN}
\begin{figure}[htb]
\begin{minipage}{0.38\columnwidth}
    \centering
    \includegraphics[width = \textwidth]{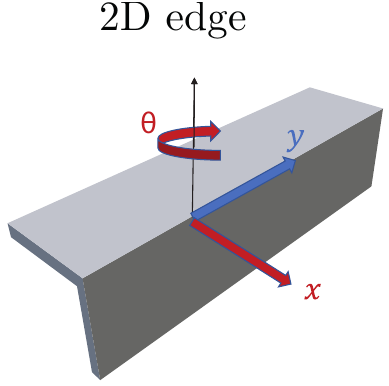}
    \label{fig:poles2}
\end{minipage}
\hfill
\begin{minipage}{0.62\columnwidth}
    \centering
\scalebox{0.80}{
\begin{tabular}{ccc}
\hline
Parameter & Range          & Shear      \\ \hline
$x$     & {[}-5, 5{]} mm    & {[}-5, 5{]} mm  \\
$y$     & -                 & {[}-5, 5{]} mm  \\
yaw, $\theta$   & {[}-45, 45{]} deg & {[}-5, 5{]} deg \\ \hline
\end{tabular}}
\end{minipage}
\caption{\small Dataset generation procedure along with its corresponding perturbations to introduce frictional shear.}
\label{fig:ranges}
 \end{figure}

%Data collection
Our sample collection was fully automated by use of a 6-DoF, IRB 120 robot arm (ABB) and TacTip tactile sensor mounted as an end effector, with software interface managed in Python and RAPID \cite{lepora2019pixels}. Tactile image data were acquired by repeatedly contacting an edge stimulus (part of the disk) across a range of poses used as labels $(x,y,\theta)$, and adding controlled unlabelled perturbations (\autoref{fig:ranges}) to generalize the shear caused by friction during the sliding motion~\cite{lepora2020optimal}. The tactile images were then cropped, thresholded and downsampled to obtain a set of $128\times128$ binary images that capture the current state of the TacTip inner pins with the corresponding pose parameters as targets. Overall, 10,000 tactile contacts were gathered over the considered pose ranges, which were split into distinct sets for training, validation and testing. For more details of the tactile sensor and data gathering framework, we refer to Ref.~\cite{lepora2019pixels}.

 \FloatBarrier
 
%Training
To estimate the contact pose $(x,y,\theta)$ from acquired tactile images, we first consider a new deterministic CNN based on the ResNet18 architecture, which was trained in PyTorch. This differs from the networks used before with this sensor~\cite{lepora2019pixels,lepora2020optimal} to enable us to implement the general uncertainty framework described above \cite{loquercio2020general}. In previous work, this network had been modified to perform regression tasks with a fully-connected layer of 8,192 hidden neurons to create an oversized structure that guarantees the redundancy needed to estimate the uncertainty. For simplicity, all outputs were normalized from -1 to 1 to prevent the sensor noise injected into the system from unilaterally affecting one of these variables. For the network optimization, the Adam algorithm \cite{kingma2014adam} was chosen, initialized with a learning rate of 0.01 and a weight decay of 0.001. Lastly, to avoid overfitting the oversized model, the L2 regularization \cite{ng2004feature} was used.

%Depending on the task, the number of outputs that can fully describe the tactile interaction is different, being 2 for 2D edge tasks and 5 for 3D ones. Thinking of the later implementation of the framework

\subsection{Uncertainty estimation with a probabilistic CNN}

\begin{figure}[htb]
    \centering
    \includegraphics[width = \columnwidth, trim = {5.2cm 9.7cm 5.2cm 3.8cm}]{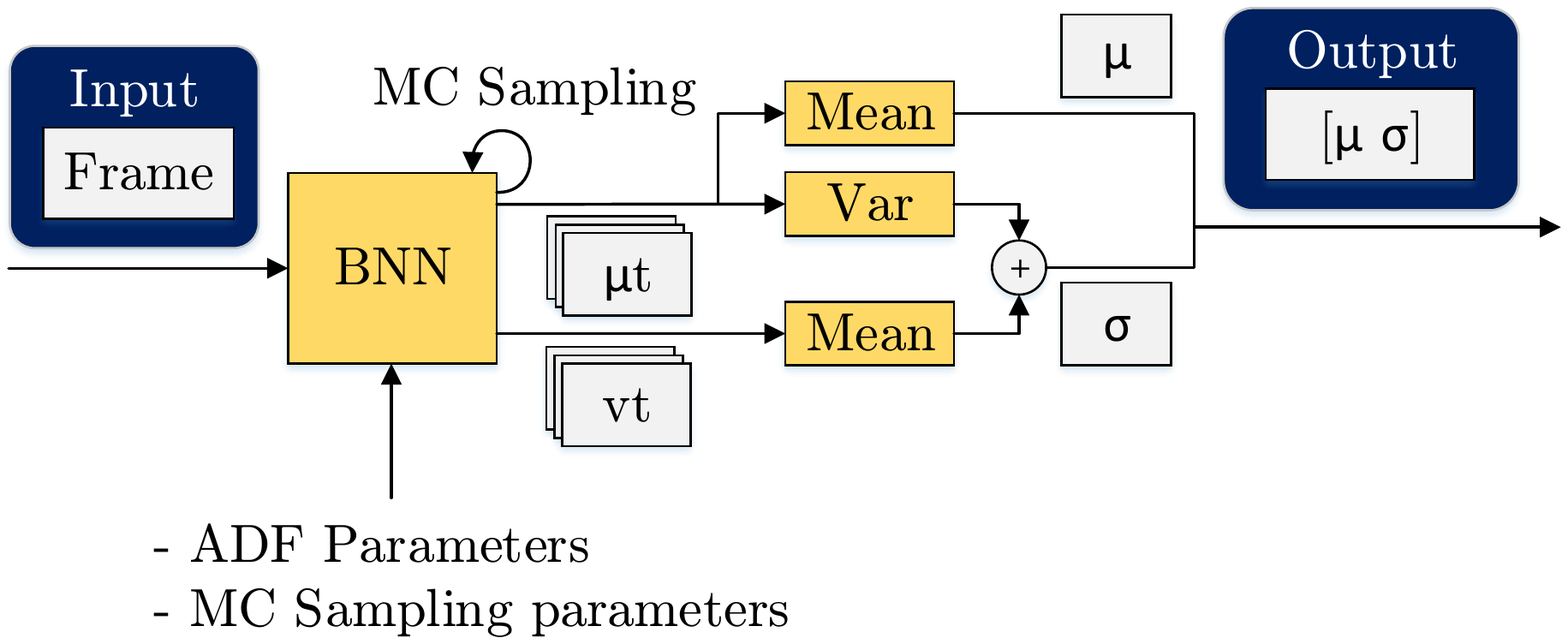}
    \caption{Overall architecture used to estimate the uncertainty proposed by \cite{loquercio2020general}. Simplified model with one single output.}
    \label{fig:uncert_framework}
\end{figure}

%Standard Uncertainty framework
Once the ResNet has been trained, the model hyperparameters and weights were loaded into its equivalent Bayesian neural network model, provided in the uncertainty framework and previously modified to have same number of hidden neurons in its fully-connected layer~\cite{loquercio2020general}. For each tactile image fed into the BNN, the architecture calculates the mean $u_t$ and variance $v_t$ of each pose parameter, according to the sensor noise. To include the model uncertainty, Monte Carlo sampling with $n$ samples is performed with dropout enabled ($p_{\rm MC}$) that results in a batch of $u_t$ and $v_t$ estimates. Therefore, when computing both sources of uncertainty, the final output can be characterized by: 1) the overall mean $\mu$ averaged over all $\mu_t$ samples; and 2) the total uncertainty $R$ from the variance of the $\mu_t$ samples and the average output uncertainty associated with the sensor noise $v_t$ (see \autoref{fig:uncert_framework}).

\FloatBarrier

%\hl{Review from here}

\begin{figure}[htb]
    \centering
    \includegraphics[width = \columnwidth]{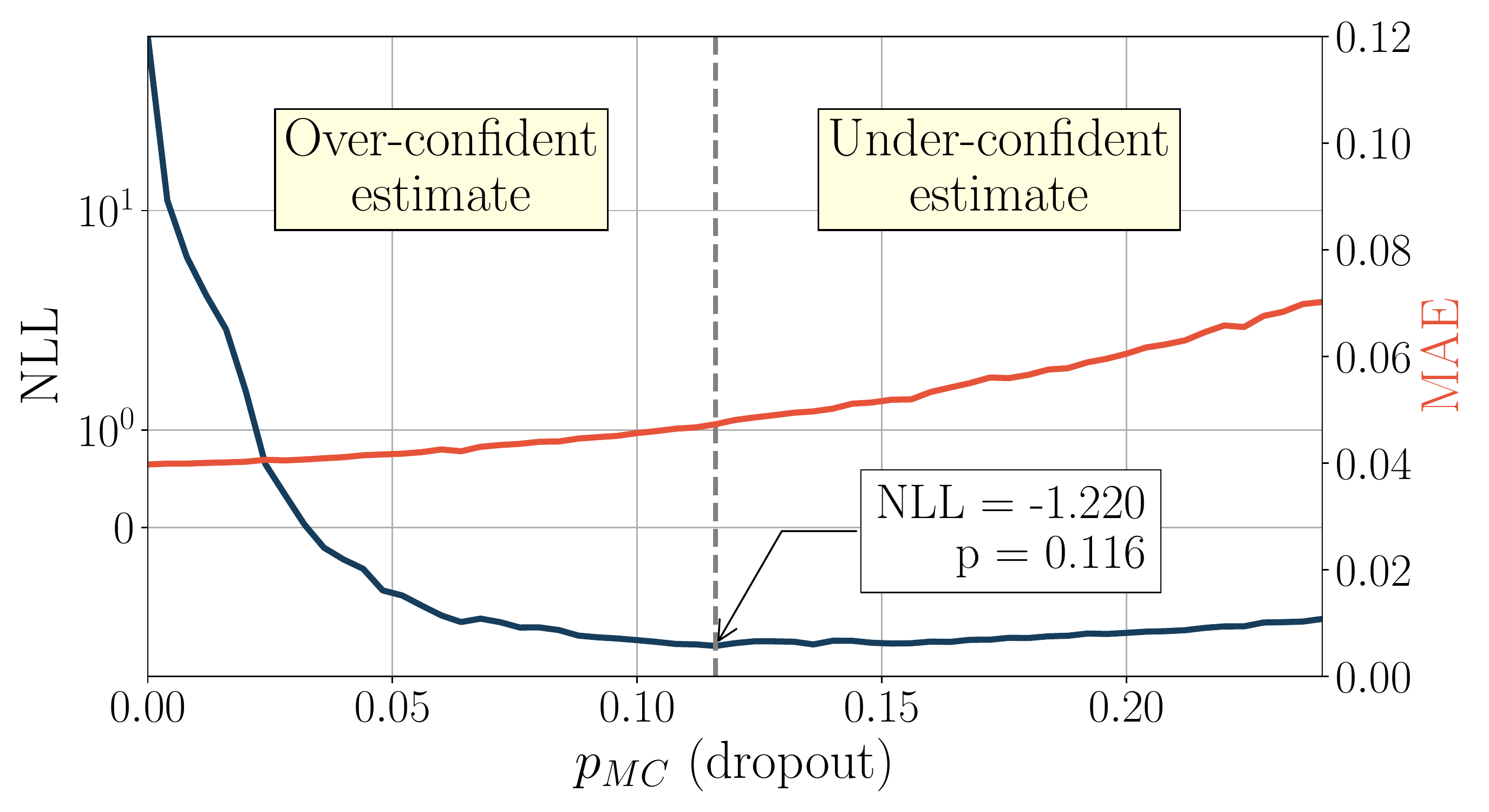}
    \caption{\small NLL and MAE metrics achieved with ResNet18 trained with $p_{\rm tr} = 0.1$ according to different $p_{\rm MC}$ dropout values used in MC sampling. Test carried out on the training dataset for the optimization of $p_{\rm MC}$. Due to the increase in deactivated neurons, the MAE tends to be higher for larger dropout rates.}
    \label{fig:min_NLL}
\end{figure}

%Optimizations
    % P_MC
The framework has two free parameters: the number of samples $n$ and the dropout probability $p_{\rm MC}$. These were chosen to optimize the quality of the uncertainty estimate according to Maximum Likelihood Estimation (MLE) applied to a Negative Log-Likelihood (NLL) metric \cite{kendall2017uncertainties} of the model success. Too small a dropout probability results in a low output variance and thus an over-confident estimate, while too high a value deactivate too many neurons, resulting in an under-confident estimate characterized by excessively high variance and mean absolute error. The $p_{\rm MC}$ value that guarantees the optimal estimate can be found by minimizing the NLL, with value typically being close to that used during training, $p_{\rm tr}$ (see \autoref{fig:min_NLL}).

%Num samples optimization
%The number of samples used ($n$) has been estimated with an iterative algorithm based on \textit{Sequential Analysis}. This method, first applied by \cite{wald1945sequential}, consists of a gradual increase of the samples to be used until a criterion set by the user is met. Thus, the user can choose to prioritise a real time computation ($\downarrow$ K) or the quality of the uncertainty estimation ($\uparrow$ K).

%The number of samples $n$ determines the balance between the computational load and the quality of the uncertainty estimation. Here we estimated the value of $n$ with a \textit{Sequential Analysis} algorithm \cite{wald1945sequential} that iteratively increases the number of samples used until a stopping criterion set by the user is met; these algorithms are known to be more sample efficient for a given accuracy than setting the sample number in advance. This way, it is possible to prioritise either a real time computation (low probability) or the quality of the uncertainty estimation (high probability).
%\begin{equation}
%\begin{aligned}
%    &{\rm NLL}_{\rm thr} = k\cdot {\rm NLL}_{\rm min}:\text{ }k\in \mathbb{R}\text{ },\text{ }0 < k < 1\\
%    &n = n_{\rm min} : {\rm NLL}(n_{\rm min}) < {\rm NLL}_{\rm thr}
%\end{aligned}
%\vspace{0.3cm}
%\end{equation}
%In our case, the uncertainty estimation was chosen to ensure the minimum number of samples were used that satisfies a NLL of at least 95\% of the minimum NLL found in the test dataset.

For the number of samples used to perform MC sampling, we use a fixed value $n=13$. This value represented a trade-off between having sufficient samples for the uncertainty estimation, while avoiding overflowing the memory of the GPU or having excessively long execution times. We checked that this value was sufficient to attain an NLL close to the minimum. 

\subsection{Bayesian filtering}
\label{sec3c}
To exploit the estimated uncertainty, we implemented a modified version of a Kalman Filter (KF) \cite{kalman1960}. The idea of this filter is to take advantage of the temporal coherence between consecutive measurements to perform better predictions. To do so, the KF algorithm calculates the optimal state of each uncorrelated pose parameter by weighting the previous state (predict step) and the predicted Gaussian distribution modelled by our architecture (measurement step). As our approach assigns a different uncertainty to each prediction, this value $R_k$ is iteratively updated during each measurement step labelled by a time index $k$:
\begin{equation}
  \begin{split}
    &Predict\text{ }step\\
    &\hat x^-_k  = \hat x_{k-1}\\
    &P^{-}_k  = P_{k-1} + Q\\
    & \\
  \end{split}
\qquad
  \begin{split}
    &Measurement\text{ }update\\
    &K_k = P^-_k(P^-_k + R_k)^{-1}\\
    &\hat x_k = \hat x^-_k + K_k(z_k - \hat x^-_k)\\
    &P_k = (1-K_k)P^-_k
  \end{split}
  \label{eq:KF}
\end{equation}

%In this equation, $\hat x^{-}_k$ and $P^{-}_{k}$ refer to the predicted state estimate and its variance, $Q$ the process variance of the filter

%\hl{prior estimated variance P-k, process variance Q, Kalman gain Kk, uncertainty of the prediction Rk.}

In the above equations, the prior estimated variance $P^{-}_k$ is updated from its previous value and the process variance $Q$, which depends only on the type of signal to be filtered. Then, the Kalman gain $K_k$ is computed from the uncertainty between both the prior variance $P^{-}_k$ and the variance $R_k$ estimated by the deep learning model. Finally, the optimal pose $\hat x_k$ and its uncertainty $P_k$ are updated by weighting the prior estimate and the output of our network $z_k$. 

Note that this Kalman filter makes a constant state assumption ($\hat x^-_k  = \hat x_{k-1}$), which is justified by our use of small step sizes during servo control. We explored using other assumptions for the predict step, but were not able to improve on our results. However, this is an interesting topic for future work that could in principle enable improved uncertainty-aware tactile servo control. 

\subsection{Contour following with tactile servo control}

\begin{figure}[h]
    \centering
    \includegraphics[width = \columnwidth]{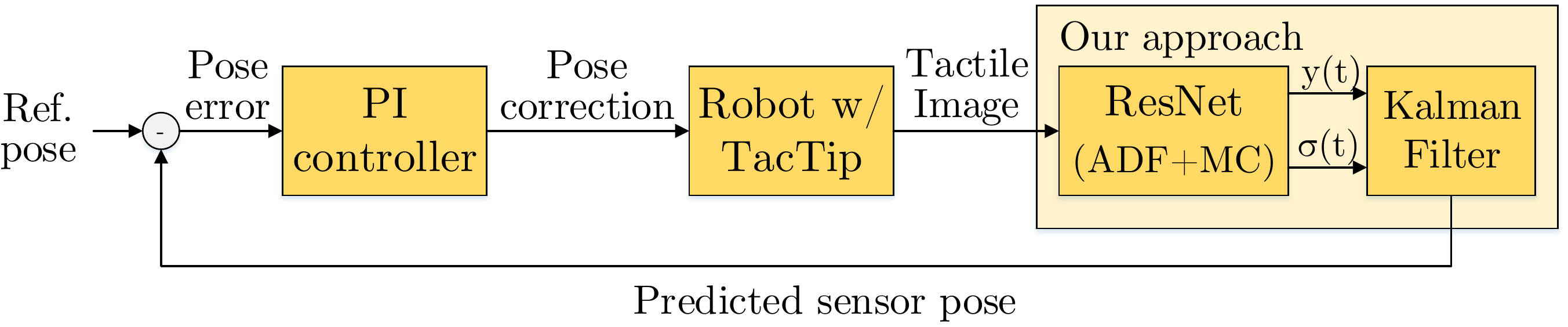}
    \caption{Probabilistic tactile servo control architecture.}
    \label{fig:my_label}
\end{figure}

To test the effectiveness of the uncertainty-aware deep learning, the Bayesian filtering was integrated into a closed-loop algorithm that allows tactile servo control with object features such as edges by maintaining a reference pose throughout the entire trajectory. The control architecture consists of 1) a PI controller, 2) a transformation to the robot base frame, 3) the tactile robot and 4) the perception module in charge of detecting changes in the pose (Figure~\ref{fig:my_label}). We refer to Ref.~\cite{lepora2019pixels} for more information about tactile servo control with this robot and the details of the tactile sensor. 

In order to establish our results, the proposed model has been tested both theoretically and practically on different co-planar shapes, including a variety of straight and curves. To measure the performance of tactile servoing, three different metrics were used, the Mean Absolute Error (MAE), Mean Squared Error (MSE) and a measure of the smoothness of the trajectory ($S_{100}$) comprised of positions ($x_i$,$y_i$):
\begin{equation*}
    \begin{aligned}
     & {\rm MAE} = \sum^{n}_{i=0} \frac{\Delta R}{n}=\sum^{n}_{i=0} \frac{|R-d\left((a,b),(x_i,y_i)\right)|}{n} \\
     & {\rm MSE} = \sum^{n}_{i=0} \frac{(\Delta R)^2}{n}=\sum^{n}_{i=0} \frac{[R-d\left((a,b),(x_i,y_i)\right)]^2}{n} \\
     & S_{100} = \frac{P^2}{A} \cdot \frac{A_{\rm est}}{P_{\rm est}^2}\cdot 100\%
    \end{aligned}
\end{equation*}
The first two error measures (MAE, MSE) are proposed for circular trajectories, with $(a,b)$ the centre of the ideal trajectory, $R$ its radius, and $d$ the Euclidean distance between two points. A third metric compatible with non-circular trajectories, $S_{100}$, is based on Cox's quantification of circularity~\cite{cox1927method}. This measure of contour irregularity was implemented to evaluate the smoothness of the trajectory from the perimeter $P$ and area $A$ of the ideal and estimated shapes.

\section{RESULTS}
\FloatBarrier
%Describe full architecture

%Describe task performed
\subsection{Offline pose estimation: Bayesian filtering}

%Note: Determinsitic performance: 0.16mm and 1.54deg

%Structure:
% Overall results are better than state-of-art methods
% Comment the errors.
% Comment the fact that limits of operating range have grater uncertainty
% KF: Comment that it is highly linear task
% Comment how MAE decreased.
% Comment how the best improvement has been achieved in MSE due to the outliers filtering stuff.

First, we assess the edge pose prediction accuracy and the uncertainty estimation by applying the probabilistic neural network to pre-collected test data. This analysis will allow us to examine how well the pose estimation model performs, so we can examine ways to improve the prediction accuracy based on the estimated uncertainty. 

\begin{figure}[t]
    \centering
    \includegraphics[width = \columnwidth]{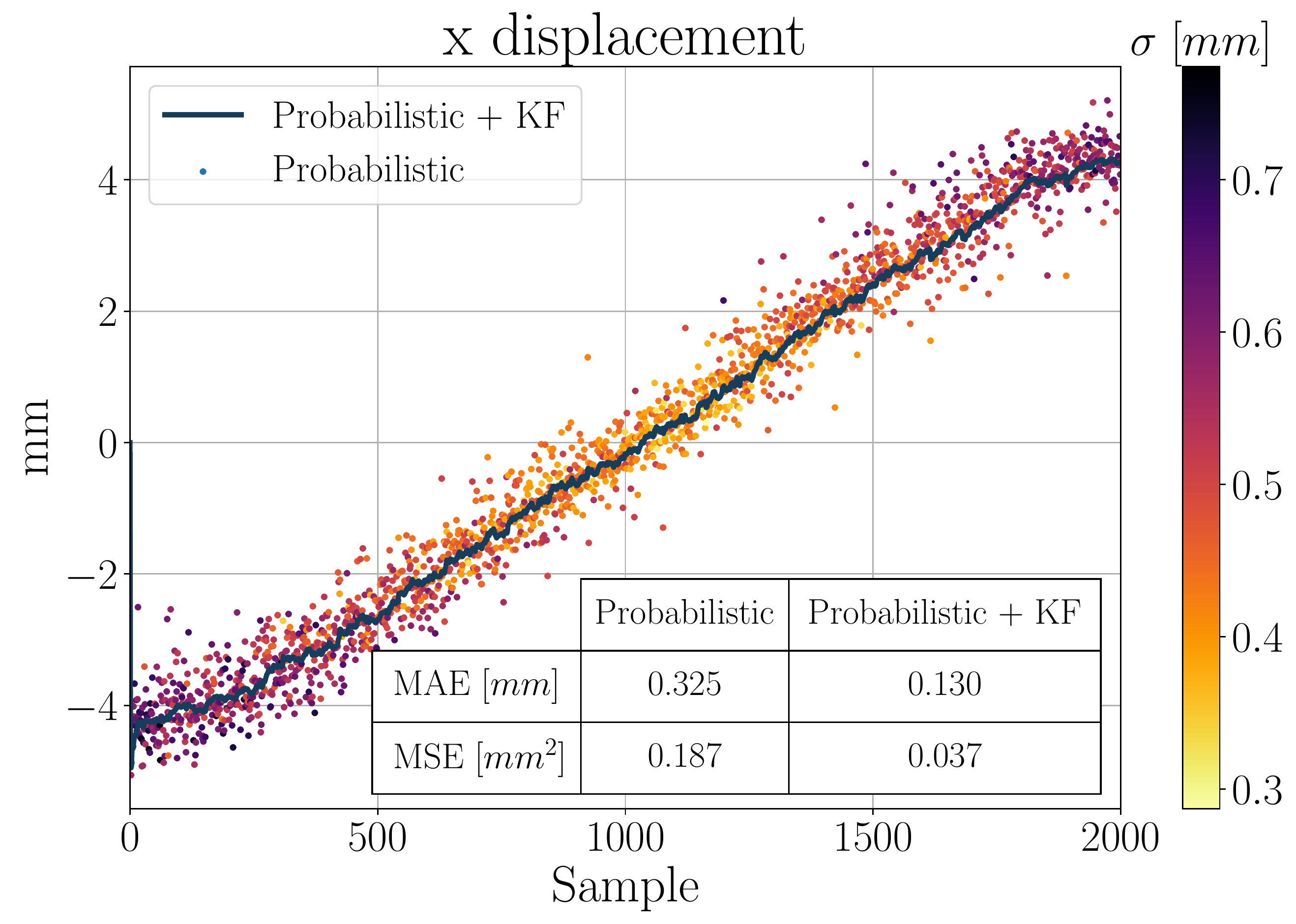}
    \vspace{0.2cm}
    \includegraphics[width = \columnwidth]{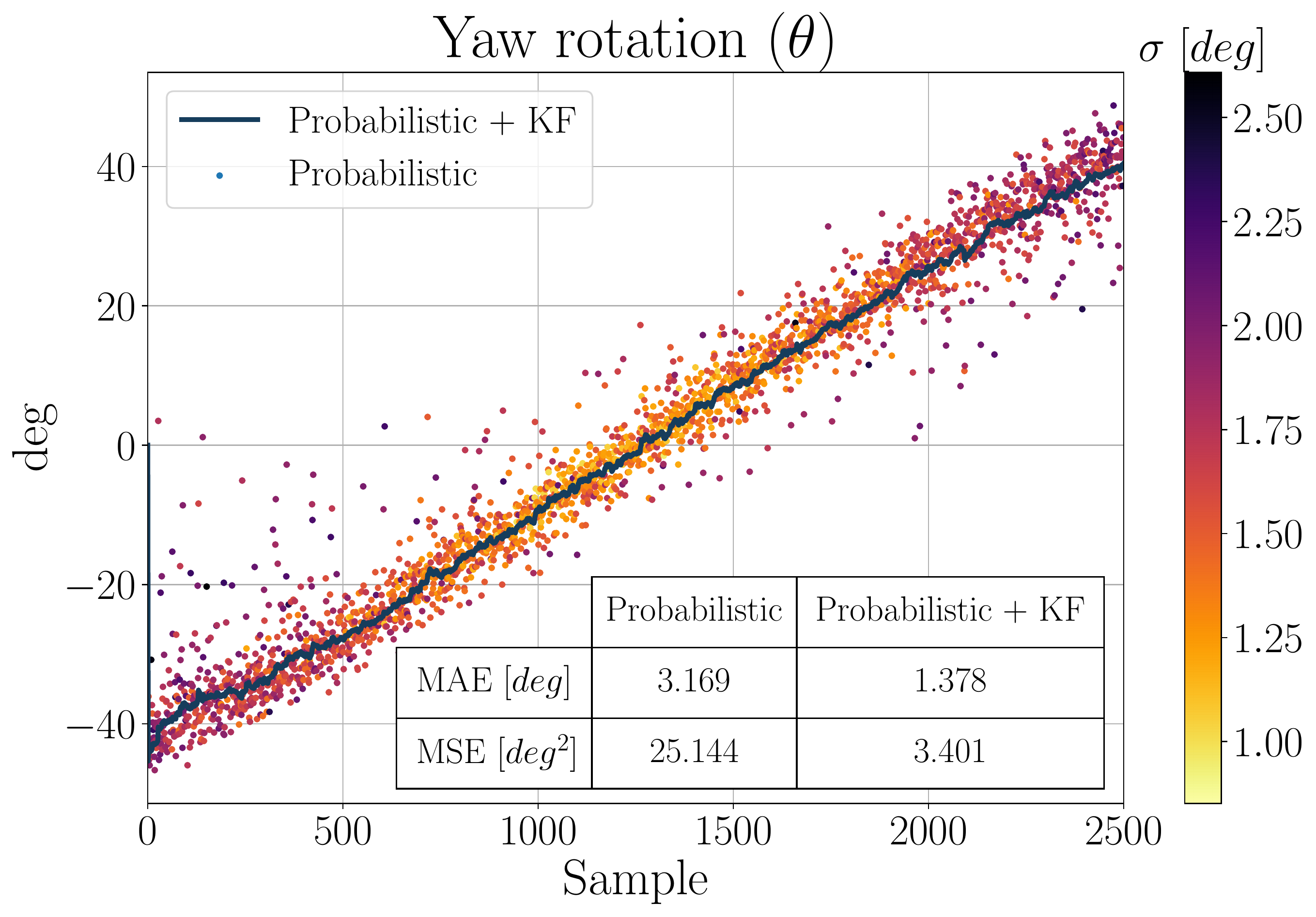}
    \caption{Performance achieved after filtering the output of the probabilistic architecture with the Kalman filter for both the $x$-displacement and $\theta$-rotation angle. Samples are ordered with increasing ground-truth label value (mm or deg).}
    \label{fig:uncert}
    %\vspace{-1em}
\end{figure}

The probabilistic ResNet18 model estimated the ($x$, $\theta$) displacement and yaw accurately to a MAE of (0.325\,mm, 3.17\,deg) respectively for individual samples (\autoref{fig:uncert}). This accuracy is similar to that found in related work with deterministic convolutional neural networks~\cite[Sec. 4A]{lepora2019pixels}. Furthermore, we assessed the effect of dropout during testing to estimate the uncertainty, to find just a small increase of (0.01\,mm, 0.1\,deg) in the overall MAE over the corresponding deterministic ResNet. 

There is a clear trend that the estimated uncertainty is smallest in the center of the parameter range at (0.3\,mm, 1\,deg), increasing to (0.75\,mm, 2.5\,deg) at the extremes of the ranges. There is also visibly more scatter at the extremes, consistent with the greater uncertainty. One explanation for this increase in uncertainty is that the regression network has less training data near the extremes; also, physically the edge becomes more ambiguous when contacted either very weakly (near free space) or on the flat top of the disk.  

To make an initial test of Bayesian filtering, these predictions were fed into a Kalman filter (\autoref{sec3c}), whose output is also shown on \autoref{fig:uncert}. As expected, the filtered predictions form a smooth curve that interpolates through the scattered data. The filtering makes these predictions far more accurate, with MAEs of (0.13\,mm, 1.39\,deg) respectively, less than half the error of the unfiltered predictions. 

Evidently, those outlying predictions that deviate from the corresponding ground truth values have a higher uncertainty that enables the filter to attenuate the effect of these wrong values. Notice also that predictions at the extremes of the ranges appear biased with respect to the ground truth towards the center of the range. A Kalman filter is not able to correct this underlying bias, only ameliorate the influence of outliers. Hence, the improvement in the centre of the range will likely be greater than that indicated by the MAEs over the entire range. This applies to the servo control examined next because the controller aims to centre the sensor in these ranges. 

%However, it is important to note that this particular task is characterised by its high linearity, which allows a better exploitation of the Kalman Filter algorithm. Depending on the application and how the pose of the TacTip changes, the efficiency of this filter might change. To overcome this problem, the number of images acquired by the camera has been increased to one frame per 0.5mm step in order to linearize the pose throughout the entire trajectory. 

\FloatBarrier

\subsection{Online tactile servo control: Contour following task}

Next, we investigated the application of uncertainty-aware pose estimation for a tactile servo control task that has been considered previously with the same experimental setup: using a sliding motion to follow the edge contours of various two-dimensional shapes~\cite{lepora2019pixels}. The robot is guided around the shapes using feedback controller to center on the edge with reference pose (0\,mm, 0\,deg), while making tangential steps $\Delta=0.5\,$mm with each iteration of the controller. Here we consider three shapes: a circular disk, clover and teardrop, with the disk appropriate for quantifying the method and the clover and teardrop demonstrating generality beyond the training set.

In general, all shapes were traced accurately, as expected from our previous work~\cite{lepora2019pixels} using convolutional neural networks. All of the generated trajectories are more accurate than in Ref.~\cite{lepora2019pixels}, both quantitatively (MAE 0.24\,mm for the disk, c.f. 1\,mm in ref.~\cite{lepora2019pixels}) and by eye from comparing the figures here to those in the previous paper. We attribute this to a smaller step size (0.5\,mm rather than 3\,mm in ref.~\cite{lepora2019pixels}), which we chose for the Kalman Filter so that the constant state assumption holds better. 

\begin{figure*}[htb]
    \centering
    \includegraphics[trim= 2cm 1cm 2.3cm 1.2cm, width = \textwidth]{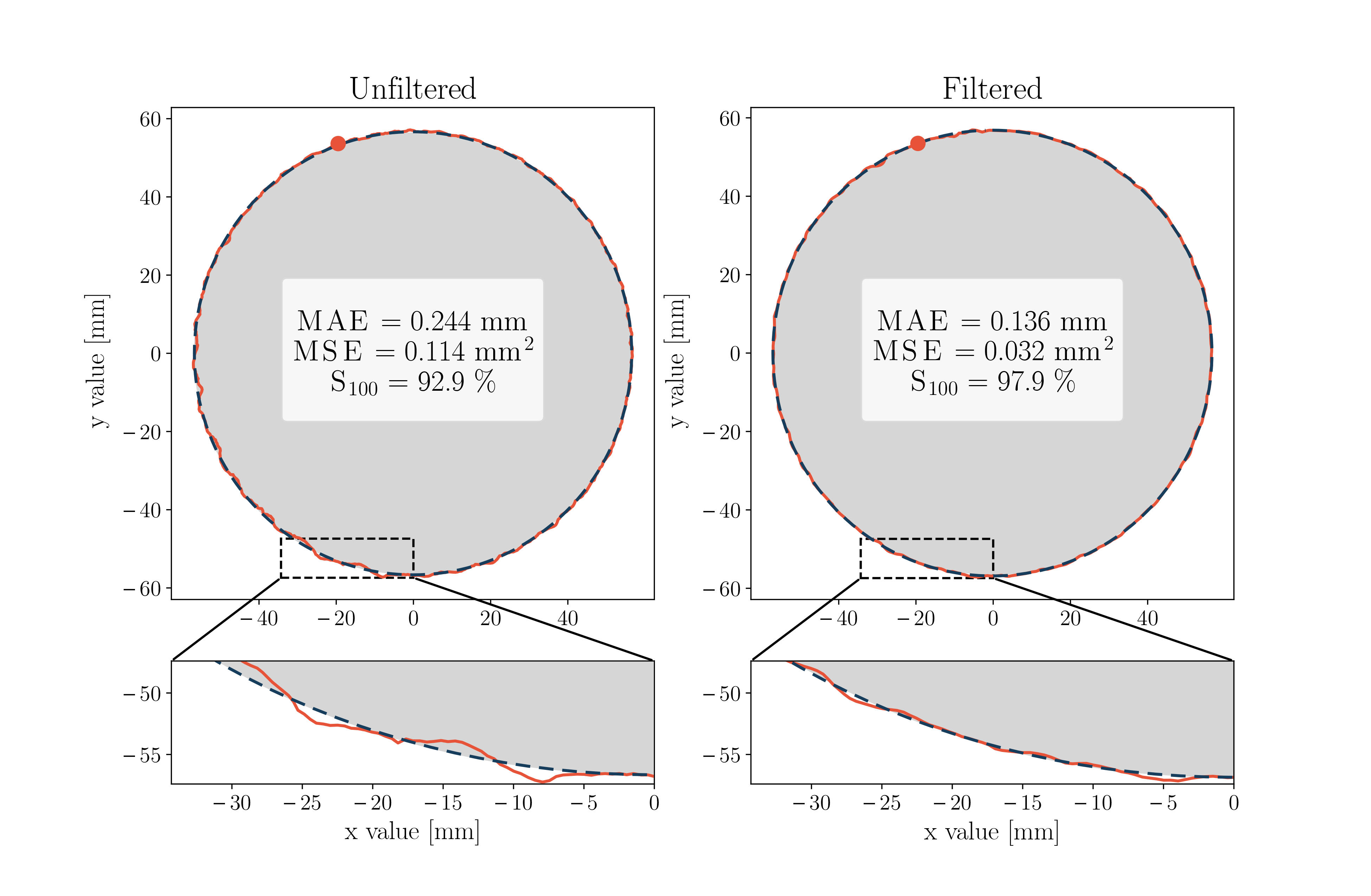}
    \caption{Circular trajectories performed by both models. $S_{100}$ represents the circularity of the trajectory from 0 to 100\%. The filtering approach was performed according to $p_{\rm MC}=0.11$ and $n=13$ MC samples. The process variance of the Kalman filter was tuned manually to 0.1\,mm for the $x$-displacement and 1\,deg for the yaw rotation.}
    \label{fig:circular}
    \vspace{-1em}
\end{figure*}

Considering the circular disk, the Kalman filtering reduces the MAE of the trajectory evaluated against a ground truth circle (from $0.24$\,mm to $0.14$\,mm), but more importantly also filters the deviations to produce a far smoother trajectory (\autoref{fig:circular}). This smoothness is quantified by the $S_{100}$ measure of circularity, showing the filtered trajectory is 2\% away from being a regular circle and the unfiltered trajectory 7\% away. The improvement in the smoothness is related to the nature of the control algorithm: as the pose parameters are kept close to zero, small deviations in the yaw rotation result can result in noisy deviations in the motion. By using a filter that exploits the temporal coherence, the inaccurate predictions are attenuated, smoothing the overall trajectory. 

%However, the introduction of BNN along with Monte Carlo sampling to estimate the uncertainty clearly affect the computing time needed. The 160ms the deterministic model needs to compute the pose from each tactile image is increased to 400ms. In spite of this, the minimum latency of each robot step is about 500ms, which makes the increment in the whole system less noticeable.

\begin{figure*}
%\vspace{0.5cm}
\centering
\begin{minipage}[b]{0.61\textwidth}
    \includegraphics[width = \textwidth]{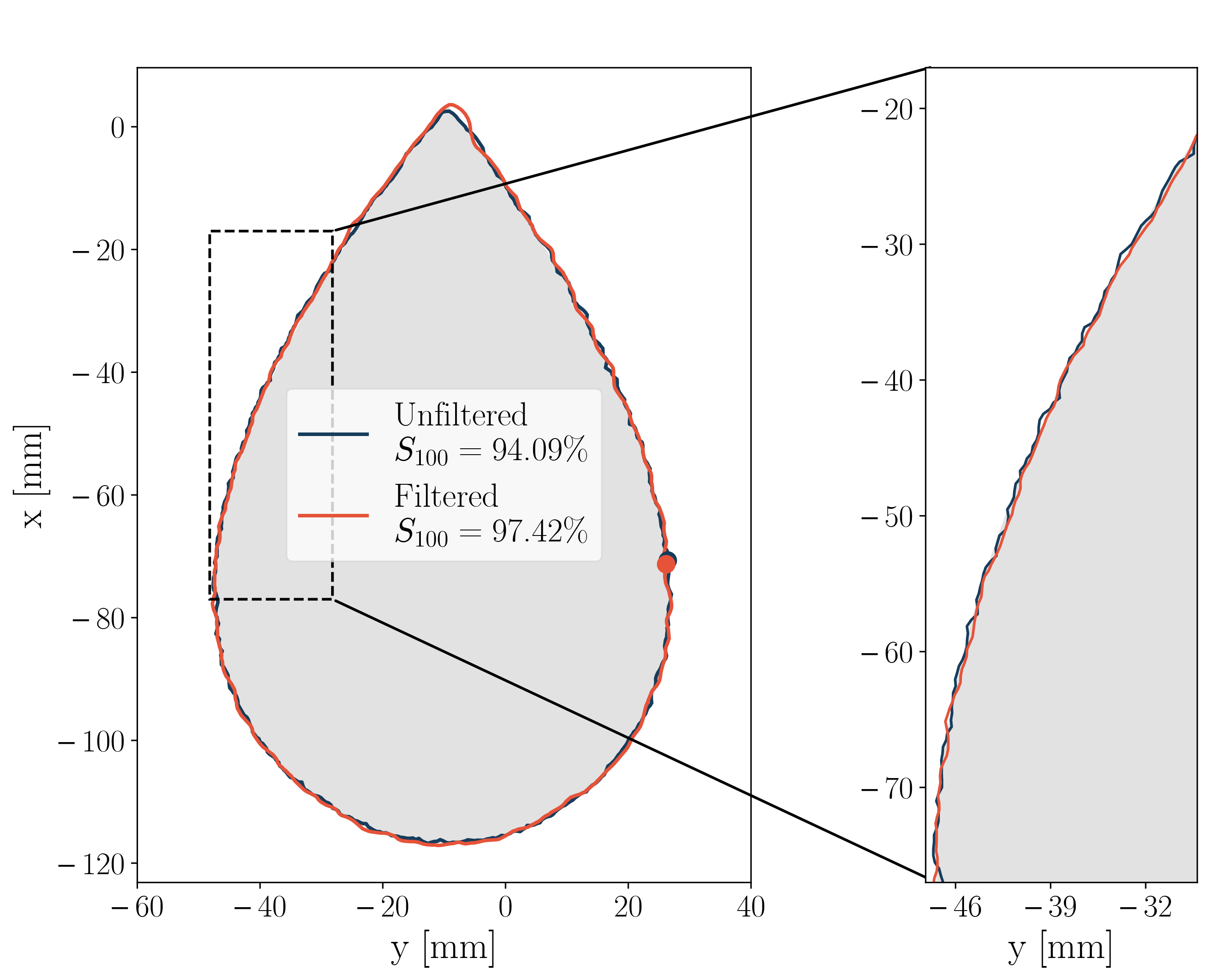}
%\caption{Caption}\label{label-a}
\end{minipage}%\qquad
\begin{minipage}[b]{0.49\textwidth}
    \includegraphics[width = \textwidth]{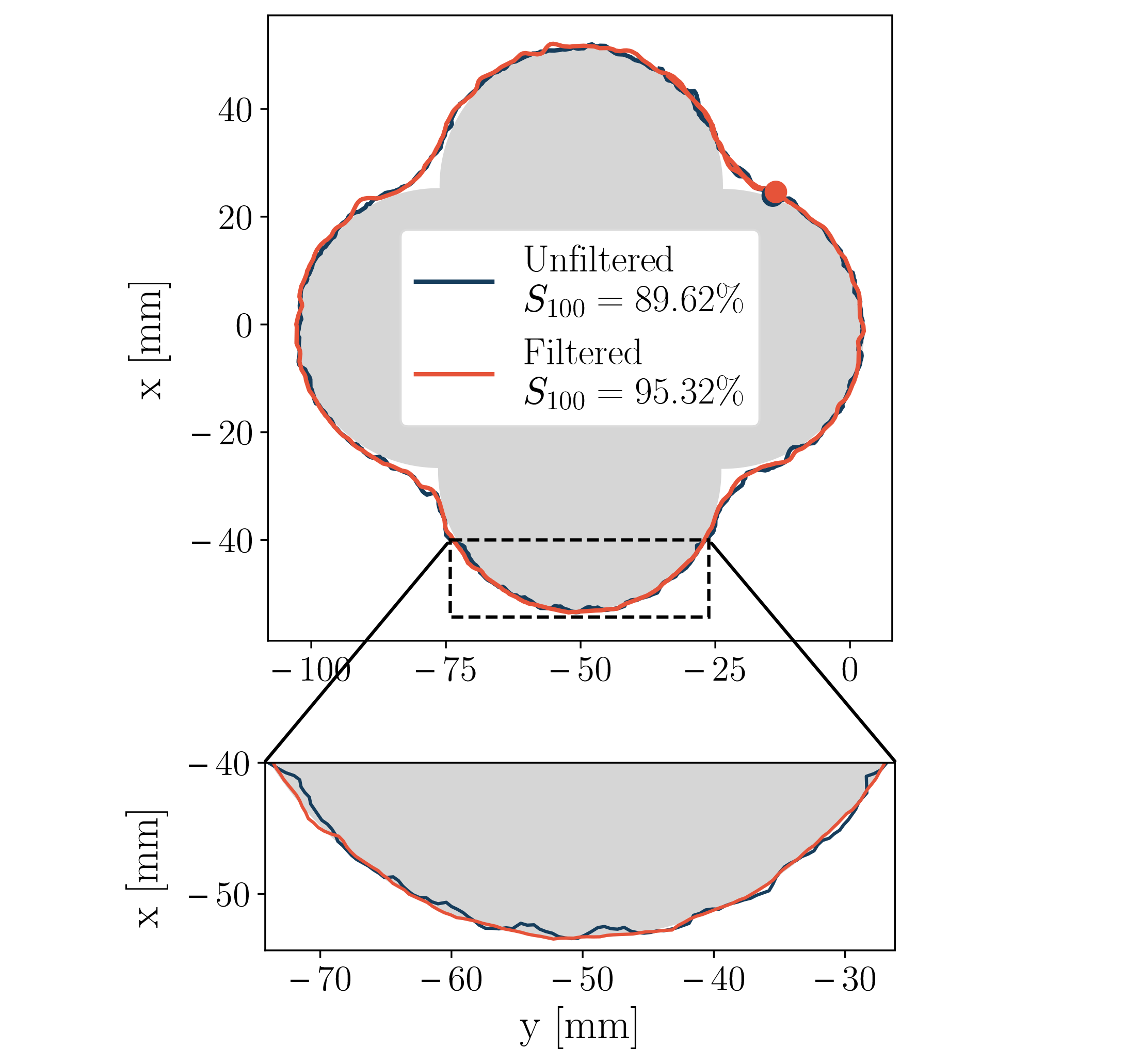}
%\caption{Caption}\label{label-b}
\end{minipage}
\caption{Non-circular shapes tested: Teardrop (left) and clover (right). Same testing conditions as Fig. 7 were used.}
\label{fig:noncircular}
\end{figure*}

% Present paragraph
% Main result: contour followed
% Differences: For teardrop-shape, comment corner case
% Differences: For clover-shape comment the difference in radiu
Tactile servoing was also performed over a teardrop and clover shapes (\autoref{fig:noncircular}) to demonstrate generality beyond the training data. Again, for both shapes the trajectories are visibly smoother after Kalman filtering, which is quantified by an improvement in the circularity $S_{100}$ that also measures the regularity of non-circular shapes. This improvement was particularly noticeable on the straight lines of the teardrop shape. However, the Kalman filter had a slower response on the sharp corner, as it attenuates sudden changes in the pose parameters. For the clover shape, the performance is similar to that achieved for the circular disk, with the filtering smoothing the overall trajectory.

% 1 column (approx)
% Restating what you have done. (succeful...)
% 2-3 main points to discuss: (connect results with rest of literature. High level)
    % Uncertainty estimation in DL (benefits: capability to improve predictions in control tasks. Observations, measurements...)
    % Temporal coherency (No RNN needed. No change in arch needed)
    % Limitations (Computing power. KF simplicty. 3D?)
    % Future opportunities 
    
\section{DISCUSSION}

%Restating what we have done
This paper developed a novel uncertainty-aware architecture for tactile servoing. By means of a recently introduced general framework for Monte Carlo sampling and Bayesian Neural Networks \cite{loquercio2020general}, the system estimates the uncertainty associated with single predictions, which is then used along with temporal coherency to improve the tactile servo control. Our results have shown how this architecture is able to deal with uncertain data to improve its filtering, which results in smoother and more precise trajectories around the edges of the various 2D objects used for testing.

% Uncertainty estimation in DL
The use of probabilistic architectures in deep learning has the potential to bring many advantages when integrated in robotic applications. The inevitable uncertainty present in robotic systems in complex environments limits the perception, so the robots will generally make sub-optimal responses. Modeling the imperfections of the sensing and the environment allows the system to be aware of this uncertainty and act accordingly. In this work, we considered Bayesian approaches to minimize this uncertainty, but there are other possibilities to control the robot to better take advantage of an uncertain environment.

% Temporal coherency
Here we used the temporal coherency present in the robot application to implement a degree of belief whose use can improve the overall performance of the system. In principle, this time feature is also present in recurrent neural network (RNN) architectures; however, their more complex training and requirement for more data makes them harder to implement. Hence, the use of Bayesian approaches with uncertainty-aware feed-forward neural networks presents a simpler and more accessible alternative. By weighting the predictions of the mathematical model and the output of the probabilistic network, these are able to predict the current state of the system more accurately. We expect this approach will apply to other applications of controlled touch.

% Limitations
A limitation of the framework is that the introduction of Monte Carlo sampling to estimate the uncertainty increases the compute time and the dedicated memory needed by the GPU. Depending on the rate of input samples to be processed, present computing resources are such that the execution of some real-time tasks may be compromised. In our case, the increase in compute time was from 160\,ms for the deterministic model to 400\,ms for the probabilistic model. As the system here only takes discrete steps about every 500\,ms, it is not greatly affected, but a key direction to progress the tactile servo control is towards more rapid cycle times. Hence, the latency from estimating the uncertainty could be an issue in future work. The continuing growth in compute power could address this issue, or more efficient deep learning networks could be implemented.   

%In terms of accuracy, we think the use of linear filters (i.e Kalman Filter) is not able to fully exploit all the advantages that a bayesian filter may offer, so we propose the exploration of more complex algorithms that allow the implementation of highly robust systems. 

% Future opportunities + ending
Estimation of uncertainty represents a first step in understanding the imperfect information in our natural surroundings. Future directions of interest include: 1) improving the post-processing of the uncertainty by implementing algorithms that result in better models (e.g. the unscented Kalman filter or particle filter), or 2) modifying the way these networks learn from uncertain information~\cite{sunderhauf2018limits}. Such architectures could weight the training samples according to their uncertainty while rejecting out-of distribution data. In our view, as the field of tactile robotics matures in its use of deep learning, the estimation of uncertainty will become a key component in the control of physically interactive robots in complex environments. 

{\em Acknowledgements:} We thank members of BRL who helped this work, including John Lloyd, Efi Psomopoulou, Kirsty Aquilina, David Barton and Ben Money-Coomes.

%The data used in this letter is available for download at \url{http://lepora.com/publications.htm} and \url{http://doi.org/bzrr}. 

\FloatBarrier

%\addtolength{\textheight}{-16cm}   % This command serves to balance the column lengths on the last page of the document manually. It shortens the textheight of the last page by a suitable amount. This command does not take effect until the next page so it should come on the page before the last.

%%%%%%%%%%%%%%%%%%%%%%%%%%%%%%%%%%%%%%%%%%%%%%%%%%%%%%%%%%%%%%%%%%%%%%%%%%%%%%%%

%\newpage
\bibliographystyle{unsrt}
\bibliography{library}

\end{document}